\documentclass[review,authoryear]{elsarticle}
\usepackage{graphicx}			
\graphicspath{{./Images/}}		
\usepackage{hyperref}			
\usepackage{pdfpages}			
\usepackage{float}				
\usepackage{gensymb}
\usepackage{amsmath}			
\usepackage{blindtext} 
\usepackage[raggedright]{titlesec} 
\usepackage{svg}
\usepackage{array} 
\usepackage{multirow} 
\usepackage{hhline} 
\usepackage{booktabs,dcolumn} 
\usepackage{fancyhdr}		
\usepackage{subcaption,booktabs} 
\usepackage{textcmds} 
\usepackage{adjustbox}
\usepackage{natbib}
\newcommand{\para}{\indent\indent}

\begin{document}

	\begin{frontmatter}
		
		\title{Semi-supervised learning for medical image classification using imbalanced training data}
		
		\author[1]{Tri Huynh\corref{cor1}}
		\cortext[cor1]{Corresponding author}
		\ead{t.huynh@latrobe.edu.au}
		
		\author[1]{Aiden Nibali}
		\author[1]{Zhen He}
		
		\address[1]{Department of Computer Science and Information Technology,
			La Trobe University, Melbourne, Australia}

		\begin{abstract}
			Medical image classification is often challenging for two reasons: a lack of labelled examples due to expensive and time-consuming annotation protocols, and imbalanced class labels due to the relative scarcity of disease-positive individuals in the wider population. Semi-supervised learning (SSL) methods exist for dealing with a lack of labels, but they generally do not address the problem of class imbalance. In this study we propose Adaptive Blended Consistency Loss (ABCL), a drop-in replacement for consistency loss in perturbation-based SSL methods. ABCL counteracts data skew by adaptively mixing the target class distribution of the consistency loss in accordance with class frequency. Our experiments with ABCL reveal improvements to unweighted average recall on two different imbalanced medical image classification datasets when compared with existing consistency losses that are not designed to counteract class imbalance.

		\end{abstract}
		
		
	\end{frontmatter}
	
	\newpage


	\pagestyle{plain}
	\section{Introduction}
\para Computer-aided diagnosis is an important field of research, as well-written algorithms can improve diagnostic efficiency and reduce the chance of misdiagnosis. For example, at the early stages tumors or lesions might be very small and easy for a radiologist to miss, but by using an algorithm to automatically highlight such instances the number of false negatives can be reduced. Recently, many supervised machine learning methods, especially those using convolutional neural networks (CNNs) have achieved promising results in medical image classification, an important component of computer-aided diagnosis ~\citep{rezvantalab2018dermatologist,han2018classification,bi2017automatic,hekler2019superior,iizuka2020deep,ausawalaithong2018automatic,rajpurkar2017chexnet}. However supervised learning requires a large number of fully labelled medical images. In the real world, annotating medical images is typically very time-consuming, especially when the consensus of multiple experts is required.  Fortunately, in many practical situations there are often many unlabelled images available that can be leveraged to boost model accuracy. Semi-supervised learning (SSL) ~\citep{chapelle2009semi}
 makes use of a small amount of labelled and a large amount of unlabelled data to  train machine learning models. The goal of SSL is to learn patterns from unlabelled data to enhance the accuracy of a model that is trained using a small amount of labelled data. There are several variants of SSL dealing with medical image classification tasks which have achieved promising results such as Categorical Generative Adversarial Networks~\citep{springenberg2015unsupervised}, Denoising Adversarial AutoEncoders~\citep{creswell2018denoising}, Dual-path Semi-supervised Conditional Generative Adversarial Networks~\citep{yang2019dscgans}, and the MeanTeacher with Label Propagation algorithm~\citep{su2019local}. Recently researchers have found perturbation based methods give the best performance~\citep{xie2019unsupervised,tarvainen2017mean,laine2016temporal,miyato2018virtual,berthelot2019mixmatch}. Hence in this study we will focus on perturbation based methods. Most perturbation based algorithms augment unlabelled data and then apply a consistency loss to make the predicted class distribution from the original and augmented (perturbed) unlabelled samples similar.
 
 There are several techniques based on this approach which recently achieved state of the art results, mostly on various classification tasks, such as Temporal Ensembling~\citep{laine2016temporal}, Mean Teacher~\citep{tarvainen2017mean}, Virtual Adversarial Training~\citep{miyato2018virtual}, Unsupervised Data Augmentation (UDA)~\citep{xie2019unsupervised} and MixMatch~\citep{berthelot2019mixmatch}.
 
Medical image datasets often have a very skewed data distribution, usually due to a large number of negative disease cases versus a small number of positive disease cases. It is critically important to address this class imbalance problem since failure to address this problem can lead to false negatives for the minor class (the disease-positive case) which can have potentially fatal consequences. This motivates us to study techniques for dealing with skewed data in SSL.	 	

Almost all existing approaches for handling skewed data distributions are only for  labelled datasets including: various oversampling or undersampling approaches ~\citep{mani2003knn,kubat1997addressing,barandela2004imbalanced,chawla2002smote,han2005borderline,bunkhumpornpat2009safe}; modified loss functions~\citep{wang2016training,lin2017focal,phan2017dnn}; cost sensitive learning~\citep{wang2018predicting,khan2017cost,zhang2016training}; decision threshold moving~\citep{krawczyk2015cost,chen2006decision,yu2016odoc}. In our case we would like to develop effective methods for addressing the skewed class distribution of unlabelled data for SSL. 

We focus our study on modifying the consistency loss (loss on unlabelled data) on one of the best performing recent SSL methods, Unsupervised Data Augmentation (UDA)~\citep{xie2019unsupervised}. Although our study is focused on UDA, our proposed method can be applied to any perturbation based SSL which uses the consistency loss such as Temporal Ensembling~\citep{laine2016temporal}, Mean Teacher~\citep{tarvainen2017mean} and Virtual Adversarial Training~\citep{miyato2018virtual}. The standard consistency loss (CL) used in UDA has two shortcomings: 1) it degrades the classification performance of the minor classes, and 2) it always sets the target as the original sample instead of a blend of the augmented and original sample. Hence, we propose the Adaptive Blended Consistency Loss (ABCL) method which tackles both shortcomings of standard consistency loss by generating a target class distribution which is a blend of the original and augmented class distributions. The blended target class distribution skews towards either the original or augmented sample depending on which predicted the minor class.

To our knowledge the only existing work that addresses the class distribution skew problem for SSL perturbation based methods is the Suppressed Consistency Loss (SCL) method~\citep{hyun2020class}. Their observation is that for class imbalanced datasets, the training is likely to move the decision boundary into the low-density areas of the minor class which will in turn cause the model to misclassify minor class samples. Therefore, SCL suppresses the consistency loss of minor classes which will help move the decision boundary away from the low-density areas that the minor classes occupy. However SCL, like the standard consistency loss of UDA, also sets its target as the original sample instead of a blend of the augmented and original samples. This will in general favor the major class since the models are more likely to predict the major class as a result of the bias inherent in the unbalanced dataset itself. In contrast, our ABCL uses a blended target class distribution that skews towards either the original or augmented sample depending on which predicts the minor class.

We have conducted an extensive experimental study on a skin cancer HAM10000 dataset~\citep{tschandl2018ham10000} and a retinal fundus glaucoma REFUGE dataset~\citep{orlando2020refuge}. Our experiments show that when our ABCL method is used to replace the consistency loss of UDA, the unweighted average recall (UAR) on HAM10000 increases from 0.59 to 0.67.  Furthermore,  ABCL also significantly outperformed the state-of-the-art SCL method for the skin cancer dataset. For the retinal fundus glaucoma dataset, ABCL significantly outperformed SCL, increasing UAR from 0.57 to 0.67. These results show that ABCL is able to improve performance of SSL for different datasets.

This paper makes the following key contributions:
\begin{itemize}
	\item We identify the importance of handling class imbalance for semi-supervised classification of medical images. In contrast, no existing work explicitly addresses this problem for medical images.
	\item We propose the Adaptive Blended Consistency Loss (ABCL) as a replacement for the consistency loss of perturbation based SSL algorithms such as UDA in order to tackle the class imbalance problem in SSL.
	\item We conduct extensive experiments on two different datasets to demonstrate the advantages of ABCL over standard consistency loss and other existing methods designed for addressing the class imbalance problem in both supervised learning and SSL. 
\end{itemize}

	\section{Material and methods}
~\subsection{Semi-supervised learning architecture}
\begin{figure}
	\centering
	\includegraphics[width=.7\textwidth]{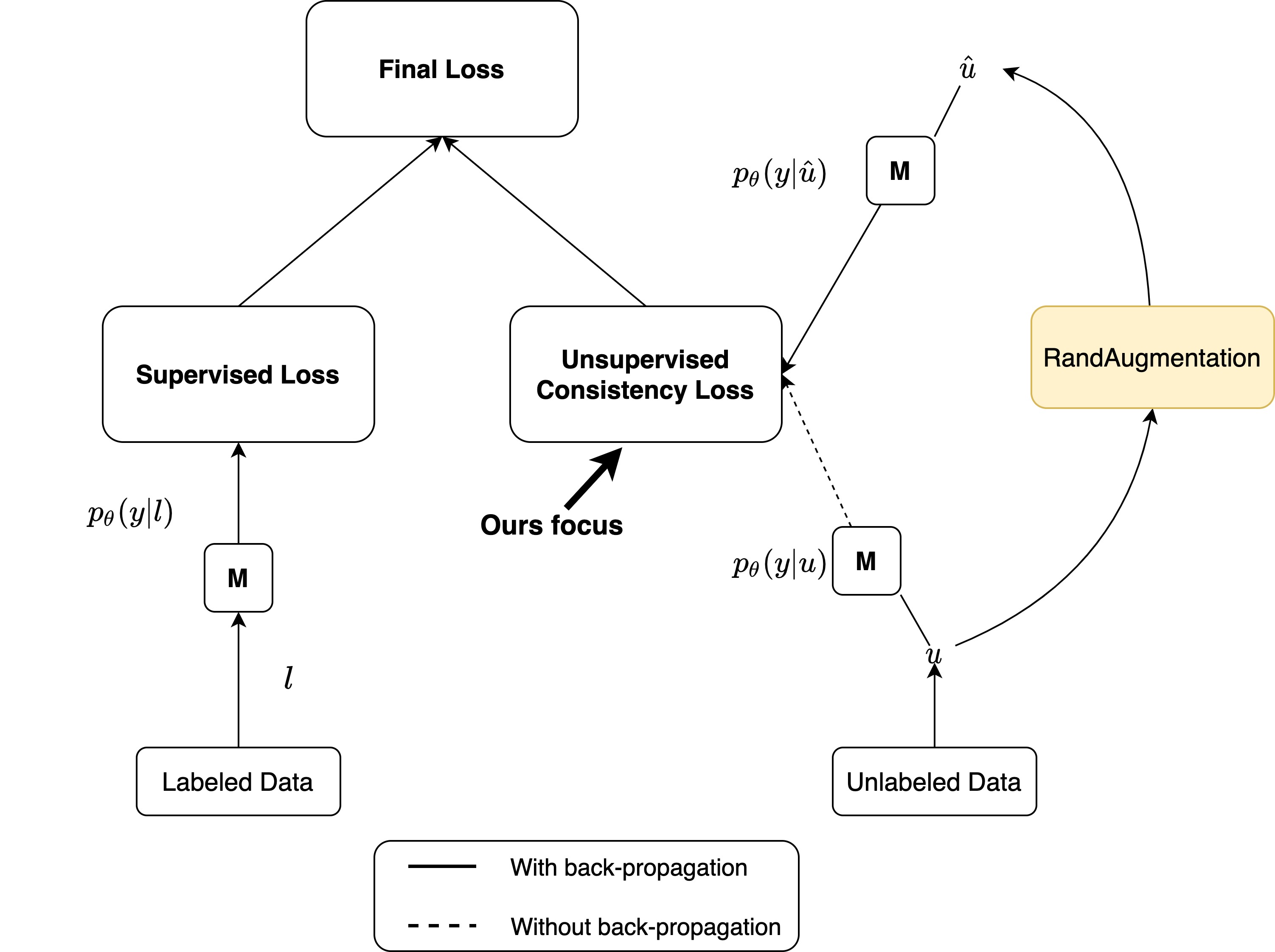}
	\caption{ The architecture used by the  Unsupervised Data Augmentation (UDA) ~\citep{xie2019unsupervised} perturbation SSL method. In the diagram M is the shared CNN model used for classifying both the labelled and unlabelled images.
	}
	\label{fig:UDA}
\end{figure}

\para We base our research on perturbation based SSL methods since all state of the art SSL methods ~\citep{xie2019unsupervised,tarvainen2017mean,laine2016temporal,miyato2018virtual,berthelot2019mixmatch} use this approach.
The solution we developed for this paper can be applied to any perturbation based SSL that uses the consistency loss. To simplify our analysis we will focus on a particular perturbation based SSL called the Unsupervised Data Augmentation~\citep{xie2019unsupervised} (UDA) method. UDA is one of the best performing recent SSL methods.~\autoref{fig:UDA} shows a diagram of how UDA works. The model is trained using two losses: a supervised loss (cross-entropy loss) and an unsupervised loss (consistency loss). The aim of consistency loss is to enforce the consistency of two prediction distributions. The key idea of UDA is to use optimal data augmentation on unlabelled samples to increase the effectiveness of the consistency loss. To obtain optimal data augmentation they applied an algorithm called RandAugmentation~\citep{cubuk2019randaugment} on the labelled dataset.
The total loss for the UDA architecture consists of two terms (see ~\autoref{fig:UDA}): supervised loss for labelled data and consistency loss for unlabelled data. The loss formula can be summarized as follows:
\begin{equation}
	\label{eqn:uda_equation}
	L =  L_{S}( p_{\theta}(y|l) ) + L_{con}( p_{\theta}(y|u) , p_{\theta}(y|\hat{u}) )
\end{equation}

Where $L_{S}$ is a supervised cross entropy loss function that takes input as the predicted probability distribution $p_{\theta}(y|l)$ of $y$ for a  labelled sample $l$ produced by the model $M$ with parameters $\theta$. $L_{con}$ is a consistency loss function that uses the Kullback-Leibler divergence to steer the predicted class distribution of the augmented unlabelled image $\hat{u}$ towards the target predicted class distribution of the original unlabelled image $u$.

A key assumption of perturbation based methods such as UDA is the smoothness assumption. Under the smoothness assumption, two data points  that are close in the input space should have the same label. That means, an unlabelled sample that is close to a labelled example will have the label information propagated to the unlabelled sample. Another important property of the smoothness assumption is that the original input and the augmented input should be close to each other in the embedded space and hence should be assigned the same label. This idea is captured by the consistency loss $L_{con}$ which ensures the model produces the same predicted probabilities from both the original and augmented input, leading to the model being robust to noise. 	

There have been various existing approaches~\citep{mani2003knn,kubat1997addressing,barandela2004imbalanced,chawla2002smote,han2005borderline,bunkhumpornpat2009safe,wang2016training,lin2017focal,phan2017dnn,wang2018predicting,khan2017cost,zhang2016training,krawczyk2015cost,chen2006decision,yu2016odoc} to tackle class imbalance for the supervised learning problem. These approaches are complementary to our solutions since they work on improving supervised loss while ours works to improve the unsupervised loss component of the overall loss. The unsupervised loss is particularly important since the number of unlabelled examples is typically much larger than labelled examples. As shown in ~\autoref{fig:UDA},  UDA uses the consistency loss to exploit the information from the unlabelled loss by making the class distribution of the augmented unlabelled data match the original unlabelled data. Hence to alleviate the class imbalance problem in UDA, we focus on modifying UDA’s consistency loss (shown in ~\autoref{eqn:uda_equation}). In particular, UDA’s consistency loss is replaced by our new novel loss function, Adaptive Blended Consistency Loss (ABCL). Hence the total loss is reformulated with ABCL replacing the unsupervised term: 
 
\begin{equation}
	L =  L_{S}( p_{\theta}(y|l) ) + ABCL( p_{\theta}(y|u) , p_{\theta}(y|\hat{u}) )
\end{equation}

\subsection{Issues with existing consistency loss formulations}
\label{sec:problems_cl_scl}
\para In this section, we analyse problems with standard consistency loss (CL) in UDA~\citep{xie2019unsupervised} and the state-of-the-art suppressed consistency loss (SCL)~\citep{hyun2020class} when training on datasets with imbalanced class distributions. The problems will be analysed in the context of the original sample’s prediction (OSP) and augmented sample’s prediction (ASP) which represent the probability distributions $ p_{\theta}(y|u)$ and  $p_{\theta}(y|\hat{u})$ respectively in ~\autoref{eqn:uda_equation}

\begin{figure}
	\centering
	\includegraphics[width=.7\textwidth]{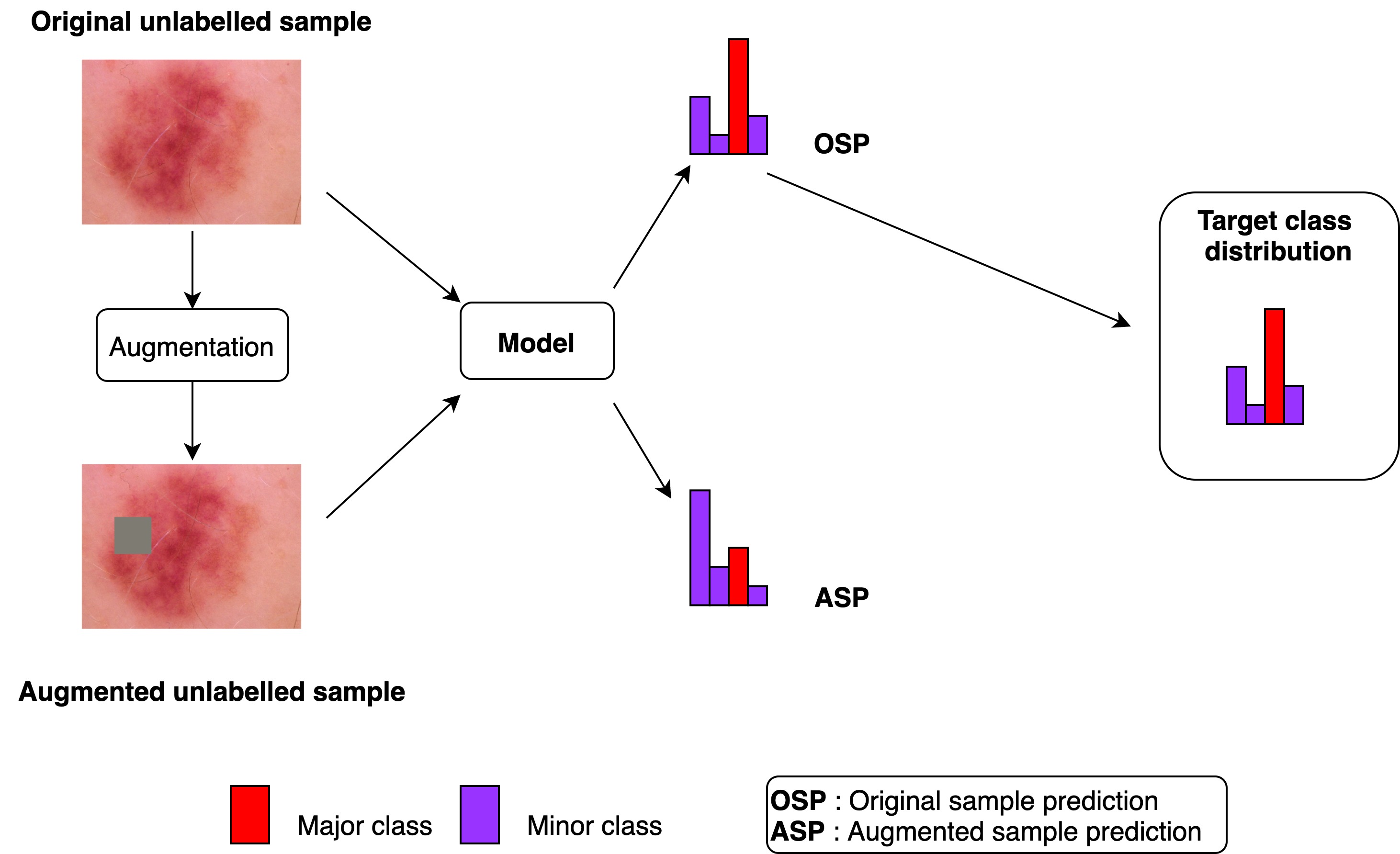}
	\caption{ An illustration showing how CL and SCL works. Note that the target class distribution is always the class distribution for OSP. In addition, SCL down-weights the loss if the predicted class of OSP is the minor class.}
	\label{fig:Normal CL}
\end{figure}

~\autoref{fig:Normal CL} illustrates how CL and SCL works. In UDA, the CL is a function that sets the OSP as the target for ASP. That is the CL always pushes the class distribution of ASP towards the class distribution of OSP. The idea behind SCL is to suppress the minor class‘s consistency loss to move the decision boundary such that it passes through a low-density region of the latent space. In practical terms, SCL suppresses the CL when the OSP is the minor class and applies the CL when the OSP is the major class. Like CL, SCL uses the OSP’s class distribution as the target irrespective of whether OSP and ASP class distributions belong to the major or minor class.

Both standard CL and SCL have two shortcomings. Firstly they are both biased towards targeting the major class in the presence of imbalanced training data. Secondly they do not target a blend of OSP and ASP but instead always target OSP only, and thus do not exploit the augmented example to improve the model's behaviour for the original example.

~\textbf{~\emph{Shortcoming 1: CL and SLC are more likely to target the major class}}. When in doubt the model will more often predict the major class since the model is trained on labelled data which is skewed towards major class samples. Consequently, samples of the minor class are more likely to be mispredicted as the major class than vice versa. In particular, when the original sample is incorrectly predicted as major class and the augmented sample is correctly predicted as minor class, the CL and SCL erroneously encourages the model to predict the sample as major class. As a result, the model’s performance will degrade for minor class samples.

~\textbf{~\emph{Shortcoming 2: CL and SLC do not target a blend of OSP and ASP but instead always targets OSP only}}. Targeting a blend of OSP and ASP reduces the harmful effects of targeting OSP only when OSP predicts the wrong class. Unlike existing methods, we do not make the assumption that the model’s prediction for the original sample is more accurate than for the augmented sample. Instead we set the target as a blend of OSP and ASP to potentially average out some of the harmful effects of a wrong OSP prediction. This is in some ways similar to the benefit of ensembling the prediction of multiple models to minimize prediction error and the established technique of test-time data augmentation~\citep{krizhevsky2012imagenet}.

The shortcomings mentioned above are all centered around what we should set as the target distribution for the CL as a function of OSP and ASP. Hence, to provide a more detailed analysis of the desired target class distribution we divide the analysis into four separate cases depending on whether major or minor classes were predicted by the OSP and ASP.
\begin{table}
	\footnotesize
	\begin{tabular}{|m{0.7cm}|m{1.7cm}|m{1.7cm}|m{1.7cm} |m{1.7cm} |m{1.7cm}|}
		\hline
		\textbf{Case} & \textbf{OSP} & \textbf{ASP} & \textbf{CL target} & \textbf{SCL target} & \textbf{ABCL target} \\
		\hline
		
		1 & Major class & Major class & OSP (major class) &OSP (major class) & Near the middle between OSP and ASP \\
		\hline
		
		2 & Minor class &Minor class &OSP (minor class) &OSP (minor class) & Near the middle between OSP and ASP \\
		\hline
		
		3 & Major class & Minor class & OSP (major class) & OSP (major class) & Closer towards ASP \\
		\hline
		
		4 & Minor class & Major class & OSP (minor class) & OSP (minor class) & Closer towards OSP \\
		\hline
	\end{tabular}
	\caption{The analysis of the target class distribution of CL (standard consistency loss), SCL (suppressed consistency loss), and our ABCL (adaptive blended consistency loss) for 4 prediction cases. ~\autoref{fig:4 case illustration} gives a diagrammatic illustration of the 4 cases.}
	\label{tab:4 cases}
\end{table}
\begin{figure}
	\centering
	\includegraphics[width=.7\textwidth]{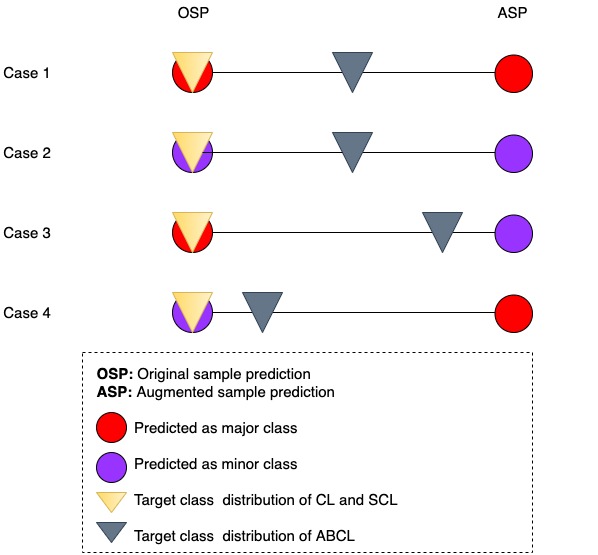}
	\caption{ An illustration of the target class distribution of CL, SCL and ABCL for the 4 cases shown in ~\autoref{tab:4 cases}.}
	\label{fig:4 case illustration}
\end{figure}
See ~\autoref{tab:4 cases} and ~\autoref{fig:4 case illustration} for an illustration of the 4 different OSP and ASP cases. In cases 1 and 2, OSP and ASP are in agreement. Unlike CL and SCL, we forgo the assumption that the OSP is in any way more valid than the ASP. As a result, we posit that it may be better to move the desired target class distribution to be a blend of OSP and ASP instead of just to OSP. This is because we would like the target to contain information in the predictions of both OSP and ASP. In case 3 CL and SCL may unintentionally encourage the bias towards the major class to be stronger since it is moving the desired target class distribution towards the major class although there is no consensus between the two predictions. There is already a natural tendency to predict the major class, a bias which is induced by the dataset skew. CL/SCL does nothing to counteract this bias as it does not consider the frequencies of the predicted classes. Case 3 presents an opportunity to counteract this bias, as the model has already indicated via the ASP that the minor class is likely correct since it made that prediction despite its natural tendency to predict the major class. Unfortunately, using CL/SCL in this situation is likely to encourage the model to mispredict the minor class sample as the major class. It would be preferable to instead move the desired target class distribution towards ASP, thus counteracting the dataset bias. In case 4, OSP is the minor class, so there is no bias towards the major class and so both CL and SCL do a good job of rewarding the minor class prediction in this case.

\subsection{Adaptive Blended Consistency Loss (ABCL)}
\label{sec:abcl}
\para To address the drawbacks of CL and SCL, the desired target class distribution for the consistency loss should be adaptively adjusted to be somewhere between OSP and ASP depending on which of the 4 cases in ~\autoref{tab:4 cases} and ~\autoref{fig:4 case illustration} has occurred. In cases 1 and 2, OSP and ASP both are either the major or minor class. In this case the desired target class distribution should be in the middle of OSP and ASP. In case 3 and 4, to discourage the bias towards predicting the major class, the desired target distribution should be closer to the minor class depending on whether OSP or ASP is the minor class. 

\begin{figure}
	\centering
	\includegraphics[width=.7\textwidth]{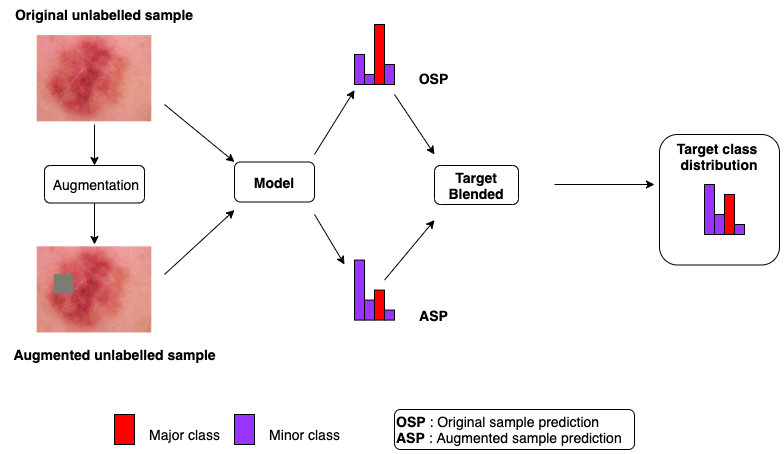}
	\caption{Diagram showing  how ABCL works. The target distribution is blended more towards the minor class, although it still retains some of OSP’s distribution.}
	\label{fig:ABCL}
\end{figure}

We proposed a new consistency loss function called Adaptive Blended Consistency Loss (ABCL) which captures the desirable properties listed above.~\autoref{fig:ABCL}  illustrates how it works. ABCL uses the following loss function to generate a new target class distribution which is a blend of OSP and ASP.
\begin{equation}
	 ABCL (z,\hat{z} ) = L_{con}(z_{blended},z) + L_{con}(z_{blended},\hat{z})
\end{equation}
\begin{equation}
	z= p_{\theta}(y|u) ; \hat{z} = p_{\theta}(y|\hat{u})  
\end{equation}
$L_{con}$ is the Kullback-Leibler divergence between the blended target probability distribution $z_{blended}$ and either OSP ($z$) or ASP ($\hat{z}$). Hence ABCL pulls both the original and augmented predictions towards the target distribution. For a model with parameters $\theta$, $p_{\theta}(y|u)$ and $p_{\theta}(y|\hat{u})$ denote the predicted class probability distribution for an original unlabelled example $u$ and its augmented version $\hat{u}$ respectively. The gradient of the loss is not be back-propagated through $z_{blended}$ during parameter optimisation.

The blended target class distribution $z_{blended}$ is defined as follows:
\begin{equation}
	z_{blended} =   (1 - k)*z+ k*\hat{z}
\end{equation}
Where $k$ is the weighting value [0,1] that determines the proportion to which the blended target moves towards OSP versus ASP. When $k$ equals 0, the target will become OSP meaning ABCL will become the same as CL. On the other hand, as $k$ approaches 1, the target approaches ASP. $k$ is calculated based on predicted class frequencies with respect to the training dataset using the following formula:
\begin{equation}
	\label{eqn:determine k}
	k= max(0, min(\gamma*(N_{original}-N_{augmented} ) + 0.5 ,1))
\end{equation}
$N_{original}$ and $N_{augmented}$ are the class frequencies of the predicted class (class with the highest predicted probability) for the OSP and ASP respectively. $\gamma \in$ (0, 1] is the class imbalance compensation strength that controls how strong the new blended target class distribution skews towards either OSP or ASP.
The value of $k$ can be interpreted as follows:
\begin{itemize}
	\item When the OSP is the minor class and the ASP is the major class (case 4 in ~\autoref{tab:4 cases} and ~\autoref{fig:4 case illustration}), the value of $N_{original}$ will be smaller than the value of $N_{augmented}$ so the value of $k$ will be smaller than 0.5. This indicates that the new blended target class distribution will skew towards the minor class side (OSP).
	\item When the OSP is the major class and the ASP is the minor class (case 3 in ~\autoref{tab:4 cases} and ~\autoref{fig:4 case illustration}), the value of $N_{original}$ will be larger than the value of $N_{augmented}$ so the value of $k$ will be larger than 0.5. This indicates that the new blended target distribution will skew towards the minor class side (ASP).
	\item When OSP and ASP both are the major class or both are the minor class (cases 1 and 2 in ~\autoref{tab:4 cases} and ~\autoref{fig:4 case illustration}), the value of $N_{original}$ and $N_{augmented}$ are similar, so the value of $k$ will be around 0.5. In this case OSP and ASP contribute equally to the blended target distribution.
\end{itemize}

~\textbf{~\emph{Class imbalance compensation strength $\gamma$}}. In ABCL (~\autoref{eqn:determine k}), the $\gamma$ term is used to control how strongly the new blended target distribution is pushed towards either OSP or ASP. As $\gamma$ approaches 0, the impact of $N_{original} - N_{augmented}$ on the value of $k$ will be smaller, and $k$ will therefore be close to 0.5. This indicates that the new blended target class distribution will skew only slightly to either OSP or ASP. On the other hand, as $\gamma$ approaches 1, the impact that $N_{original} - N_{augmented}$ has on $k$ will be bigger. This indicates that the new blended target class distribution will skew strongly to either the OSP or ASP. When $\gamma$ is set to a high value, ABCL approaches the standard CL in the case that $N_{augmented} >> N_{original}$, since $z_{blended} \approx z$. This corresponds to CL, where OSP is wholly responsible for defining the target of the unsupervised loss term.

	\section{Experiment Setup}
\subsection{Dataset}
\begin{figure}
	\centering
	\subfloat[]{\includegraphics[width=1\textwidth]{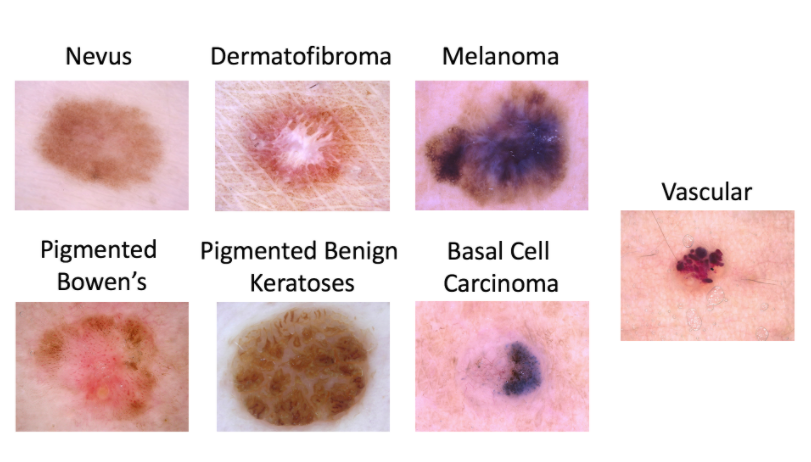}} 
	\newline
	\subfloat[]{\includegraphics[width=1\textwidth]{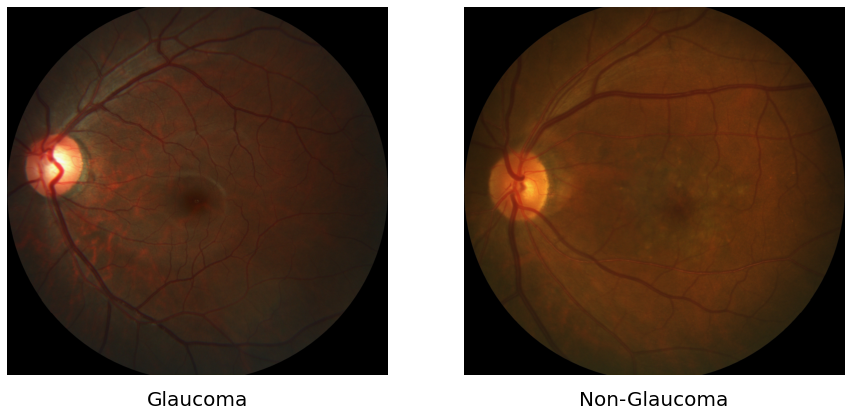}} 
	\caption{Example images of the dermatoscopic skin lesions and retinal fundus glaucoma dataset for each class.}
	\label{fig:classificationimage}
\end{figure}

\para We evaluate our proposed methodology on skin cancer and retinal fundus glaucoma datasets:

\begin{itemize}
	\item HAM10000 ~\citep{tschandl2018ham10000} (main dataset) : contains 10015 RGB dermatoscopic skin lesions images of size (450, 600) multiclass, divided into 7 classes: Pigmented Bowen’s (AKIEC), Basal Cell Carcinoma (BCC), Pigmented Benign Keratoses (BKL), Dermatofibroma (DF), Melanoma (MEL), Nevus (NV), Vascula (VASC).
	\item REFUGE Challenge~\citep{orlando2020refuge} (supporting dataset) : contains 1200 RGB retinal fundus glaucoma images of various sizes. Each image is assigned a binary label of glaucoma or non-glaucoma.
	
\end{itemize}
~\autoref{fig:classificationimage} shows example images for each class. As is shown in ~\autoref{tab:classificationdistribution}, the class distributions of both datasets are highly imbalanced.
\begin{table}
	\begin{subtable}{1\textwidth}
		\scriptsize
		\begin{tabular}{| l | l | l | l | l | l | l | l | l | l |}
			\hline
			\multicolumn{1}{|c|}{\multirow{3}{*}{\textbf{Skin cancer dataset}}}  &\textbf{MEL} &\textbf{NV} &\textbf{BCC} &\textbf{AKIEC} &\textbf{BKL} &\textbf{DF} &\textbf{VASC} &\textbf{Total} \\
			\cline{2-9}
			& 1113  & 6705  & 514  & 327  & 1099  & 115  & 142 & 10015 \\
			\cline{2-9}
			& 0.11 & 0.67 & 0.06 & 0.03 & 0.11 & 0.01 & 0.01 & 1 \\
			\hline
			
		\end{tabular}
	\end{subtable}
	\bigskip
	\newline
	\vspace*{0.5 cm}
	\begin{subtable}{1\textwidth}
		\scriptsize
		\begin{tabular}{| l | l | l | l |}
			\hline
			\multicolumn{1}{|c|}{\multirow{3}{*}{\textbf{Retinal fundus glaucoma dataset}}}  & \textbf{Glaucoma} &\textbf{Non-Glaucoma)} &\textbf{Total}  \\
			\cline{2-4}
			& 120  & 1080 & 1200  \\
			\cline{2-4}
			& 0.1 & 0.9 & 1 \\
			\hline
		\end{tabular}
	\end{subtable}
	\caption{The class distribution of 2 experimental classification datasets. Both datasets are very imbalanced. The range of class frequencies frequency is (0.67,0.01) for skin cancer dataset and (0.9,0.1) for retinal fundus glaucoma dataset.}
	\label{tab:classificationdistribution}
\end{table}

We split the datasets as follows: 70$\%$ training images, 20$\%$ test images and 10$\%$ validation images. We use the validation dataset to decide when to perform early stopping when training the model. All the results reported in the paper are from the test dataset. Notably, the unlabelled training dataset is deliberately larger than the labelled training dataset in order to simulate the  problem we described earlier. We ensure the class distribution across each of the data splits are the same. 

\subsection{Data processing}
\para For the HAM10000 dataset, the model is trained with the original image size of (450, 600), in contrast, images of the REFUGE challenge dataset are resized to (512,512). All images of both datasets have their color normalised. In particular, the red and blue color channel will be normalised based on the green color channel. The formula is described as follows: 	
\begin{equation*} 
       R = R * ( \frac{G_{mean}}{R_{mean}}) ; 	B = B *(\frac{G_{mean}}{B_{mean}})      
\end{equation*}

\subsection{Evaluation metrics}
\para Due to the class imbalance in the testing dataset, the model usually performs really well on the major classes and much worse on the minor classes. By looking at the overall accuracy, we might be tempted to say that the model has good performance, but actually it only achieves high accuracy on major classes. However, minor class accuracy is really important, especially for medical image analysis since often the samples with disease belong to the minor class. Hence, the Skin Lesion Analysis towards Melanoma Detection (ISIC 2018)~\citep{gutman2016skin} competition used an evaluation metric that gives equal weight to all classes (called unweighted average recall (UAR)) to rank the performance of the different models. Therefore, we also use UAR to measure our model’s performance across all classes in a fair way. In addition we report results using the G-mean and average Area Under the ROC curve (AUC) metric since they are also commonly used in other studies involving classification of imbalanced classes~\citep{kaur2019systematic}.   	

\subsection{Model training}

\para All models were trained for 200 epochs with the static learning rate of $10^{-4}$. To implement early stopping we use the model that gave the best validation UAR for measuring results on the test set. Furthermore, the Stochastic Gradient Descent (SGD) optimizer with momentum 0.9 and weight decay 5 x $10^{-4}$ was used. During training we use a batch size of 30 samples consisting of 8 labelled samples and 22 unlabelled samples.

\subsection{Data Augmentation}
\label{sec:data_augmentation}
\para Strong data augmentation via RandAugmentation~\citep{cubuk2019randaugment} is a key component of the UDA method, and the authors show that it can significantly improve the sample efficiency by encouraging consistency on a diverse set of augmented examples. However, experiments presented in ~\autoref{sec:weak_strong} show ABCL performs poorly when strong data augmentation is used. We attribute the poor performance of ABCL in the presence of strong data augmentation to increased distance between the augmented and unaugmented examples in embedding space. ABCL assumes in cases 3 and 4 of ~\autoref{tab:4 cases} (one of the predictions belongs to a major class and the other belongs to a minor class), it is more likely the actual class is the minor class. However, when using strong data augmentation, the augmented major class sample may be moved far away from the original example in feature space. As a consequence, there is a risk the augmented sample may be predicted to be a different class. That is, samples belonging to the major classes are more likely to be predicted wrongly since the strong augmentation can make very significant changes to the appearance of the images. Consequently, the main assumption behind ABCL is violated when strong data augmentation is used. Hence, we use weak data augmentation for both labelled and unlabelled samples which include random horizontal flips, rotations (between 0 and 180 degree), erasing~\citep{zhong2017random} (a random proportion (0.02-0.33) of input image will be erased) and color augmentation (jitter brightness, saturation and contrast with a random value in range of 0.9 to 1.1).

\subsection{Algorithms used in experimental study}

\para \textbf{\emph{UDA baseline}}. The original UDA method is implemented to obtain the baseline method. This baseline is not designed to handle class imbalance. All other methods implemented modify this baseline UDA method. As default we use weak data augmentation for all algorithms implemented in this paper including UDA baseline. However in~\autoref{sec:weak_strong} we show an experiment where we compare the performance of strong versus weak data augmentation for UDA and UDA+ABCL. The results show that ABCL with weak data augmentation is able to outperform UDA using strong data augmentation. This shows that benefits of using ABCL outweigh the performance loss from using weak data augmentation. 

\textbf{\emph{Sampling based method for the labelled dataset}}. We will resample the skewed labelled dataset by using the intelligent oversampling method SMOTE~\citep{chawla2002smote} and random undersampling.

\textbf{\emph{Weighted Cross Entropy Loss (WeightedCE)}}. This is an extension of the Cross Entropy loss function which weights the loss of each class based on a given set of weights. The objective of WeightedCE is to give higher loss for the minor class and lower loss for the major class. Intuitively, it helps the model reduce the effect of the major class. We define the set of class weights as follows:
\begin{itemize}
	\item Class frequency :  [0.11,0.67,0.06,0.03,0.11,0.01,0.01] for the HAM10000 dataset and [0.1,0.9] for the REFUGE dataset
	\item weight = 1 - Class frequency 
\end{itemize}

\textbf{\emph{Focal Loss~\citep{lin2017focal}}}. This loss function is used for supervised learning to reduce the effect of the \qq{easy samples} (samples belonging to the major class). The default $\gamma$ value used was 1 because we found this value for $\gamma$ gave the best UAR result when we did a hyperparameter search.

\textbf{\emph{Consistency loss suppression (SCL)~\citep{hyun2020class}}}. This is the state-of-the-art SSL method used to address class skew that suppresses the consistency loss when the minor class is predicted. We implemented this approach on top of UDA.  The default $\beta$ value used was 0.8 because we found this value gave the best UAR result when we did a hyperparameter search.

\textbf{\emph{Adaptive Blended Consistency Loss (ABCL)}}. This is our method for addressing the class imbalance problem for SSL implemented on top of UDA. Our new method has only one hyperparameter: the class imbalance compensation factor $\gamma$. The default $\gamma$ value used was 0.4 because we found this value gave the best overall result when considering UAR and the individual recall results for each class. 

\textbf{\emph{UDA-WeightedCE-SCL and UDA-WeightedCE-ABCL}}. This method combines supervised and unsupervised class imbalance methods. We combine WeightedCE with SCL and ABCL. We chose WeightedCE since we found WeightedCE performs the best among all the supervised class imbalance methods.

	\section{Experiment Results}
~\subsection{Results for the  skin cancer (HAM10000) dataset}
\label{sec:skin_resullt}
~\autoref{tab:overall_results} shows the test set results (UAR as the main metric and G-mean, average AUC as supporting metrics) comparing our ABCL with other methods for the HAM10000 dataset. All semi-supervised methods are based on the UDA method. Our ABCL achieved the highest UAR, G-mean and average AUC values of 0.67, 0.62 and 0.95 respectively among all experimented methods. Existing class imbalance methods designed for labelled data (Sampling and Focal) performed worse than the baseline UDA method for the UAR and G-mean metric meaning they were not effective at addressing the class imbalance problem in SSL. This is because these methods can only make a small contribution to overall model quality since they can only be applied to the labelled data which in SSL is just a small portion of the total dataset. 
On the other hand, class imbalance methods designed for unlabelled data (SCL and our ABCL) outperformed the baseline UDA for the UAR and G-mean metrics meaning these methods are more effective than methods which are only applicable to labelled data. This is because these methods modify the consistency loss which is used for the unlabelled data (occupies a larger proportion of all data for SSL) .

When WeightedCE was used with ABCL and SCL, the UAR performance was boosted from 0.67 to 0.68 and from 0.61 to 0.65 respectively. This can be explained by the fact  addressing class imbalance for the labelled data helps the model produce lower bias of the major class for the pseudo target distribution in the unsupervised consistency loss. 
Additionally, although the SCL outperformed the baseline UDA, its performance was still worse than our ABCL in all experiments that involved both methods. This is because as discussed in ~\autoref{sec:problems_cl_scl}, SCL has the two problems of bias towards major class and ignoring the augmented sample prediction when determining the target class distribution.

To further understand the performance of ABCL against the rival methods, we also reported the test set recall of each class in ~\autoref{tab:recall_classes}. This allows us to see how the major and minor classes contribute to the UAR. ABCL performed the best among all methods for almost all of the minor classes (MEL,BCC,BKL,VASC) and performed worse than most methods for the major class NV. This result shows ABCL is doing what it was designed to do, namely, it alleviates the bias towards the major class unlike CL and SCL.

\begin{table}
	\begin{tabular}{l c c c}
		\toprule
		\textbf{Algorithms} & \textbf{UAR} & \textbf{G-mean}   & \textbf{Average AUC}\\
		\midrule
		Supervised learning & 0.75 & 0.71 & 0.98\\
	
		UDA baseline & 0.59 & 0.56 & 0.91\\
	
		UDA-Sampling & 0.51 & 0.44 & 0.89\\ 
	
		UDA-WeightedCE & 0.59 & 0.55 & 0.92\\
		
		UDA-Focal  & 0.55 & 0.50 & 0.92\\
		
		UDA-SCL & 0.61 & 0.58 & 0.92\\
		
		UDA-WeightedCE-SCL & 0.65 & 0.64 & 0.94 \\
		UDA-ABCL(ours) & 0.67& 0.62 & 0.95\\
		UDA-WeightedCE-ABCL(ours) & \textbf{0.68} & \textbf{0.66} & \textbf{0.96} \\
		\bottomrule
	\end{tabular}
	\caption{ Test set results of supervised learning, UDA baseline and various methods designed for handling class imbalance on top of UDA for the HAM10000 dataset. The results in bold show the best result for each column among the unsupervised methods.}
	\label{tab:overall_results}
\end{table}

\begin{table}
	\scriptsize
	\begin{tabular}{m{2cm} m{0.7cm} m{0.7cm} m{0.7cm} m{0.9cm} m{0.7cm} m{0.7cm} m{0.7cm} | m{0.7cm}}
		\toprule
		\textbf{Algorithms} &\textbf{MEL (0.11)} &\textbf{NV (0.67)} &\textbf{BCC (0.06)} &\textbf{AKIEC (0.03)} &\textbf{BKL (0.11)} &\textbf{DF (0.01)} &\textbf{VASC (0.01)} &\textbf{UAR} \\
		\midrule
		UDA baseline & 0.43 & 0.93 & 0.81 & 0.35 & 0.55 & 0.50 & 0.56 & 0.59 \\
		
		UDA-Sampling & 0.44 & 0.84 & 0.69 & 0.52 & 0.53 & 0.36 & 0.12 & 0.51 \\ 
		
		UDA-Weighted Loss & 0.39 & 0.94 & \textbf{0.83} & 0.40 & 0.53 & 0.41 & 0.59 & 0.59 \\ 
		
		UDA-Focal & 0.34 &\textbf{0.96} & 0.81 & 0.32 & 0.51 & 0.32 & 0.62 & 0.55 \\ 
		
		UDA-SCL & 0.40 & 0.94 & 0.78 & 0.46 & 0.57 & 0.45 & 0.65 & 0.61 \\
		
		UDA-ABCL(ours) & \textbf{0.73} & 0.88 & \textbf{0.83} & 0.46 & 0.64 & 0.36 & \textbf{0.76} & 0.67 \\
		
		UDA-WeightedCE-SCL & 0.5 & \textbf{0.96} & 0.73 & \textbf{0.58} & 0.59 & \textbf{0.55} & 0.65 & 0.65 \\
		UDA-WeightedCE-ABCL(ours) & 0.68 & 0.9 & 0.79 & 0.51 & \textbf{0.67} & 0.45 & 0.74 & \textbf{0.68} \\
		
		\bottomrule
		\hline
	\end{tabular}
	\caption{ Test set recall results of the algorithms for each class of the HAM10000 dataset. The number in brackets under each class name shows the fraction of all samples belonging to that class. Therefore the major class with most examples is NV.}
	\label{tab:recall_classes}
\end{table}

~\textbf{\emph{The importance of high recall for minor classes.}} In the medical domain it is usually better to mistakenly report a false positive than to miss a true positive disease diagnosis. This is because manual assessment or further testing can be used to correct the misdiagnosis. However, missing a positive disease diagnosis may leave a potentially fatal condition untreated. Additionally, in medical data, the amount of healthy cases is usually higher than the amount of diseased cases implying the major class is usually the healthy case. In the skin cancer dataset, the NV class is the benign class, which is also the only major class and the rest are minor classes. Based on this principle, ABCL gives better performance than all its competitors. The test set results in ~\autoref{tab:recall_classes} shows that ABCL when compared to its nearest non-ABCL rival achieves higher recall for the minor classes MEL, BCC, BKL and VASC. This means ABCL is less likely to miss disease diagnosis than alternative losses.
\begin{figure}
	\centering
	\includegraphics[width=1\textwidth]{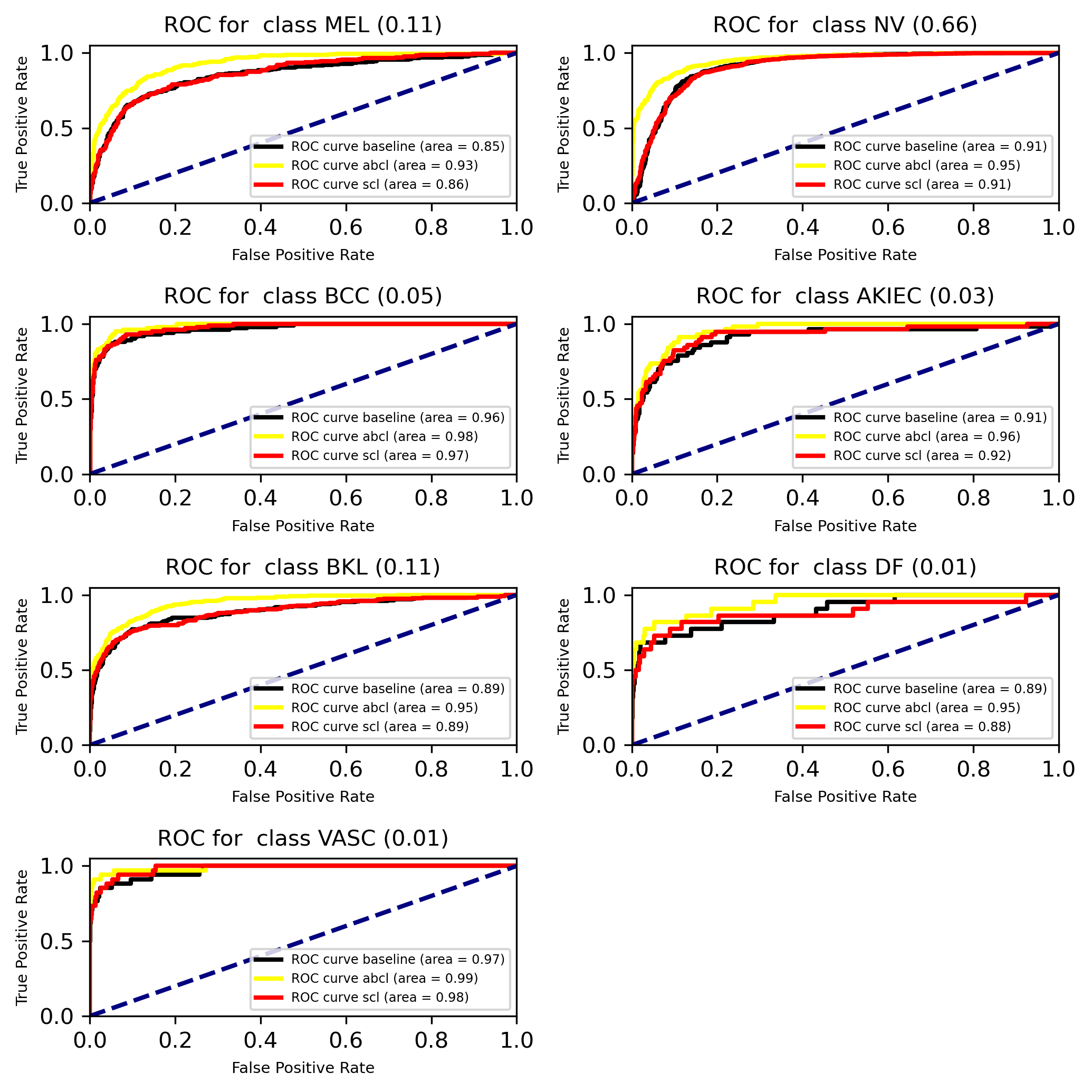}
	\caption{Test set ROC and AUC results for each class of the HAM10000 dataset using UDA baseline, SCL and ABCL methods. The number in the brackets next to the class name is the fraction of examples that belong to that class.}
	\label{fig:ROC}
\end{figure}

\begin{table}
	\scriptsize
	\begin{tabular}{ m{1.6cm} m{0.7cm}  m{0.7cm} m{0.7cm} m{0.8cm} m{0.7cm} m{0.7cm} m{0.7cm} | m{0.9cm}  }
		\toprule
		\textbf{Algorithms} &\textbf{MEL (0.11)} &\textbf{NV (0.67)} &\textbf{BCC (0.06)} &\textbf{AKIEC (0.03)} &\textbf{BKL (0.11)} &\textbf{DF (0.01)} &\textbf{VASC (0.01)} &\textbf{Average AUC} \\
		\midrule
		UDA baseline & 0.85 & 0.91 & 0.96 & 0.91 & 0.89 & 0.89 & 0.97 & 0.91 \\
		
		UDA-ABCL(ours) & \textbf{0.93} & \textbf{0.95} & \textbf{0.98} & \textbf{0.96} &\textbf{0.95} & \textbf{0.95} & \textbf{0.99} & \textbf{0.95} \\
		
		UDA-SCL& 0.86& 0.91 & 0.97 & 0.92 & 0.89 & 0.88 & 0.98 & 0.92 \\ 
		
		\bottomrule
		\hline
	\end{tabular}
	\caption{ Test set AUC results for each class of the HAM10000 dataset and its average using UDA baseline, SCL and ABCL methods. The number in the brackets next to the class name is the fraction of examples that belong to that class.}
	\label{tab:auc}
\end{table}
~\autoref{fig:ROC} shows our ABCL method compared against the SCL and UDA baseline using the ROC and the corresponding AUC. ABCL’s AUC results are better than the UDA baseline and SCL for all classes according to ~\autoref{tab:auc}. This means ABCL is better at separating between the positive and negative classes than the alternatives.

\subsection{The effects of varying $\gamma$ value}
\para In this section we analyse the important $\gamma$ parameter of ABCL according to reported recall results for each class in ~\autoref{tab:recall_gamma}. Important observations include the following:
\begin{itemize}
	\item As $\gamma$ approaches 1, the recall of the major class NV  decreases and the recall of almost all minor classes increases.
	\item As $\gamma$ approaches 0, the recall of the major class NV  increases and the recall of almost all minor classes decreases.
\end{itemize}

This implies that as $\gamma$ is increasing, the model is compensating more towards the minor classes than the major class, leading to the degradation of major class performance. This can be explained by the case when the model correctly predicts the OSP as the major class and mispredicts the ASP as the minor class, the blended target class distribution is then skewed towards the ASP. As $\gamma$ approaches 1, the blended target class distribution moves closer to ASP. Therefore the $\gamma$ term in ABCL can be used to tradeoff decreased major class performance for increased minor class performance. 

From ~\autoref{tab:recall_gamma} we decided to choose a $\gamma$ value of 0.4 as our default value since it provided a good balance of recall for the minor classes while retaining most of the recall for the major class NV.

\begin{table}
	\scriptsize
	\begin{tabular}{m{0.8cm} m{0.8cm} m{0.8cm} m{0.8cm} m{0.8cm} m{0.8cm} m{0.8cm} m{0.8cm} | m{0.8cm}}
		\toprule
		\textbf{$\gamma$ value} &\textbf{MEL (0.11)} &\textbf{NV (0.67)} &\textbf{BCC (0.06)} &\textbf{AKIEC (0.03)} &\textbf{BKL (0.11)} &\textbf{DF (0.01)} &\textbf{VASC (0.01)} &\textbf{UAR} \\
		\midrule
		0.2 & 0.66 & \textbf{0.92} & 0.82 & 0.47 & 0.65 & 0.32 & 0.68 & 0.65 \\
		
		0.4 & 0.73 & 0.88 & 0.83 &0.46 & 0.64 & 0.36 & 0.76 & 0.67 \\ 
		
		0.5 & 0.74 & 0.86 & 0.84 & 0.51 & 0.64 & 0.41 & 0.71 & 0.67 \\ 
		
		0.6 & 0.73 &0.83 & 0.84 & 0.51 & \textbf{0.66} & 0.36 & 0.76 & 0.67 \\ 
		
		0.8 & 0.75 & 0.78 & \textbf{0.86} & 0.49 & 0.63 & \textbf{0.45} &0.76 & 0.68 \\
		
		1 & \textbf{0.78} & 0.76 & 0.83 & \textbf{0.53} & 0.64 & 0.41 & \textbf{0.82} & \textbf{0.68} \\
		\bottomrule
		\hline
	\end{tabular}
	\caption{Test set recall results for each class of the HAM10000 dataset when the $\gamma$ hyper parameter value of ABCL is varied.}
	\label{tab:recall_gamma}
\end{table}

\subsection{Ablation study}
\subsubsection{Selective versus always-on target blending}
\para Here we compare ABCL with selective target blending versus ABCL with always-on target blending using the HAM10000 dataset. In our experiments we use ABCL with always-on target blending as our default method. This means even when the original and augmented samples both are predicted as the minor or major class (cases 1 and 2 of ~\autoref{tab:4 cases}), we still blend the two samples to produce the targets. However in selective target blending we do not blend the original and augmented predictions when they both predict the minor or major class, instead in this situation we resort to the standard UDA method of just setting the target as the predicted output from the original sample.

The results in ~\autoref{tab:recall_blend_select} show that the version of ABCL that always blends targets gives higher UAR (between 0.65 and 0.68) across the entire range of $\gamma$ values. In contrast, ABCL with selective target blending performs poorly for high $\gamma$ values, especially for the major class NV. ABCL improves the model’s performance on minor classes by compensating more towards minor classes. As a consequence, in case 2 of ~\autoref{tab:4 cases} (the original and augmented samples both are predicted as the minor class), there is a harmful effect to major classes that the actual major class might be mispredicted as the minor class. Therefore, in this case, always selecting the original distribution as the target distribution might exacerbate this harmful effect. On the other hand, ABCL with always-on blending can mitigate this harmful effect, leading to less degradation in the recall for the major class.

\begin{table}
	\scriptsize
	\begin{tabular}{ m{1.4cm} m{0.7cm} m{0.7cm} m{0.7cm} m{0.7cm} m{0.8cm} m{0.7cm} m{0.7cm} m{0.7cm}  | m{0.7cm}}
		\toprule
		\textbf{Algorithms} & \textbf{$\gamma$ value} &\textbf{MEL (0.11)} &\textbf{NV (0.67)} &\textbf{BCC (0.06)} &\textbf{AKIEC (0.03)} &\textbf{BKL (0.11)} &\textbf{DF (0.01)} &\textbf{VASC (0.01)} &\textbf{UAR} \\
		\midrule
		
		\multicolumn{1}{c  }{\multirow{6}{0.7cm}{ ABCL (always-on blending)} }  & 0.2 & 0.66 & \textbf{0.92} & 0.82 & 0.47 & 0.65 & 0.32 & 0.68 & 0.65 \\
		
		\multicolumn{1}{ c  }{}  & 0.4 & 0.73 & 0.88 & 0.83 &0.46 & 0.64 & 0.36 & 0.76 & 0.67 \\ 
		
		\multicolumn{1}{ c }{}  & 0.5 & 0.74 & 0.86 & 0.84 & 0.51 & 0.64 & 0.41 & 0.71 & 0.67 \\ 
		
		\multicolumn{1}{ c }{}  & 0.6 & 0.73 &0.83 & 0.84 & 0.51 & \textbf{0.66} & 0.36 & 0.76 & 0.67 \\ 
		
		\multicolumn{1}{ c }{}  & 0.8 & 0.75 & 0.78 & \textbf{0.86} & 0.49 & 0.63 & 0.45 &0.76 & 0.68 \\
		
		\multicolumn{1}{ c }{} & 1 & \textbf{0.78} & 0.76 & 0.83 & 0.53 & 0.64 & 0.41 & \textbf{0.82} & \textbf{0.68} \\
		
		\midrule
		
		\multicolumn{1}{c  }{\multirow{6}{0.7cm}{ ABCL (selective blending)} }  & 0.2 & 0.52 & 0.91 & 0.77 & 0.49 & 0.60 & 0.45 & 0.65 & 0.63 \\
		
		\multicolumn{1}{ c  }{}  & 0.4 & 0.65 & 0.83 & 0.83 &0.49 & 0.62 & \textbf{0.55} & 0.74 & 0.67 \\ 
		
		\multicolumn{1}{ c }{}  & 0.5 & 0.64 & 0.80 & 0.85 & 0.50 & 0.61 & 0.53 & 0.72 & 0.66 \\ 
		
		\multicolumn{1}{ c }{}  & 0.6 & 0.65 &0.69 & 0.85 & 0.46 & 0.55 & 0.45 & 0.76 & 0.63 \\ 
		
		\multicolumn{1}{ c }{}  & 0.8 & 0.69 & 0.53 & \textbf{0.86} & 0.53 & 0.58 & 0.32 &0.76 & 0.61 \\
		
		\multicolumn{1}{ c }{} & 1 & 0.59 & 0.50 & 0.82 & \textbf{0.56} & 0.59 & 0.41 & 0.71 & 0.60 \\
		
		\bottomrule
	\end{tabular}
	\caption{ Test set recall results selective target blending versus always-on blending for each class of the HAM10000 dataset. }
	\label{tab:recall_blend_select}
\end{table}

\subsubsection{Weak versus strong data augmentation}
\label{sec:weak_strong}
\begin{table}
	\begin{tabular}{c c c c}
		\toprule
		\textbf{Algorithms} & \textbf{UAR} & \textbf{G-mean} & \textbf{Average AUC}\\ 
		\midrule
		UDA baseline + WeakAug & 0.59 & 0.56 & 0.91 \\
		UDA baseline + StrongAug & 0.61 & 0.56 & 0.92 \\
		ABCL + WeakAug & \textbf{0.67} & \textbf{0.62} & \textbf{0.95} \\
		ABCL + StrongAug & 0.50 & 0.42 & 0.91 \\
		\bottomrule
		
	\end{tabular}
	\caption{ Test set UAR, G-mean and average AUC results of the HAM10000 dataset comparing UDA baseline and ABCL with strong data augmentation and weak data augmentation.}
	\label{tab:strongweak}
\end{table}
\para In this section we compare the effects of different data augmentation strategies for the baseline UDA method and our ABCL method on the HAM10000 dataset.~\autoref{tab:strongweak} shows the results of this experiment. The results show that for the UAR, G-mean and average AUC metrics, strong data augmentation works better than weak data augmentation for the UDA baseline whereas the opposite is true for ABCL. As discussed in ~\autoref{sec:data_augmentation}, we believe the reason for this is when using strong data augmentation, even samples belonging to major classes may be predicted wrongly since the augmentation can make very significant changes to the appearance of the images. Consequently, the main assumption behind ABCL is violated when strong data augmentation is used. The results also show that ABCL with weak data augmentation is able to outperform the baseline UDA using the strong data augmentation. This is an important result since it shows ABCL is able to overcome any negative consequence of not being able to use strong data augmentation.

\subsection{Results for the  retinal fundus glaucoma (REFUGE) dataset}
\begin{table}
	\begin{tabular}{l c c c}
		\toprule
		\textbf{Algorithms} & \textbf{UAR} & \textbf{G-mean}   & \textbf{Average AUC}\\
		\midrule
		UDA baseline & 0.55 & 0.37 & 0.82\\
		
		UDA-WeightedCE & 0.57 & 0.43 & 0.83\\
		
		UDA-Focal  & 0.57 & 0.43 & 0.82\\

		UDA-SCL & 0.55 & 0.37 & 0.81\\
		
		UDA-WeightedCE-SCL & 0.55 & 0.37 & 0.82 \\
		UDA-ABCL(ours) & 0.55& 0.37 & 0.82\\
		UDA-WeightedCE-ABCL(ours) & \textbf{0.67} & \textbf{0.61} & \textbf{0.83} \\
		\bottomrule
	\end{tabular}
	\caption{ Test set UAR, G-mean  and AUC results for the REFUGE dataset.}
	\label{tab:refuge_result}
\end{table}
\para ~\autoref{tab:refuge_result} shows the test set results of comparing ABCL with the other competing methods for the glaucoma (REFUGE) dataset. ABCL does not improve the UDA-baseline model’s performance when the standard cross entropy loss is used for the supervised loss. However, using a combination of WeightedCE for the supervised loss and ABCL for unsupervised loss, UDA-WeightedCE-ABCL is able to significantly outperform UDA-WeightedCE and UDA-WeightedCE-SCL. As explained in ~\autoref{sec:skin_resullt}, WeightedCE can help the model produce lower bias of towards the major class for the pseudo target distribution in the unsupervised consistency loss which is of benefit to both ABCL and SCL methods. However, UDA-WeightedCE-ABCL outperforms UDA-WeightedCE-SCL because ABCL is able to adaptively blend the target distribution between the original and augmented samples’s predicted distributions depending on which predicted the minor class. This helps ABCL better address the class imbalance problem in the unlabelled data.

	\section{Conclusion}
\para In this study, we identified an important gap in existing literature: class imbalance in the context of semi-supervised learning for medical images. This is an important problem to study since medical image datasets often have skewed distributions and missing positive disease diagnosis can have fatal consequences. We address this gap by proposing a new consistency loss function called Adaptive Blended Consistency Loss (ABCL). To demonstrate the effectiveness of ABCL, we applied it to the perturbation based semi-supervised learning algorithm UDA. ABCL extends standard consistency loss by generating a new blended target class distribution from the mix of original and augmented sample’s class distribution. Extensive experiments showed ABCL consistently outperforms alternative approaches designed to address the class imbalance problem. Our proposed ABCL method was able to improve the performance of UDA baseline from 0.59 to 0.67 UAR, outperform methods that address the class imbalance problem for labelled data (between 0.51 and 0.59 UAR) and for unlabelled data (0.61 UAR) on the imbalanced skin cancer dataset. On the imbalanced retinal fundus glaucoma dataset, combining with Weighted Cross Entropy loss, ABCL achieved 0.67 UAR as compared to 0.57 to its nearest rival. The results on both datasets show that ABCL can be combined with methods designed to address class imbalance for supervised learning to further boost performance on semi-supervised learning. As future work we plan to evaluate ABCL on other medical imaging tasks, such as semi-supervised semantic segmentation.

	~\section*{Declaration of Competing Interest}
\para The authors declare that they have no known competing financial interests or personal relationships that could have appeared to influence the work reported in this paper.
	\section*{Acknowledgements}
\para This research did not receive any specific grant from funding agencies in the public, commercial, or not-for-profit sectors.

	\bibliography{references.bib}

\begin{thebibliography}{40}
\expandafter\ifx\csname natexlab\endcsname\relax\def\natexlab#1{#1}\fi
\providecommand{\url}[1]{\texttt{#1}}
\providecommand{\href}[2]{#2}
\providecommand{\path}[1]{#1}
\providecommand{\DOIprefix}{doi:}
\providecommand{\ArXivprefix}{arXiv:}
\providecommand{\URLprefix}{URL: }
\providecommand{\Pubmedprefix}{pmid:}
\providecommand{\doi}[1]{\href{http://dx.doi.org/#1}{\path{#1}}}
\providecommand{\Pubmed}[1]{\href{pmid:#1}{\path{#1}}}
\providecommand{\bibinfo}[2]{#2}
\ifx\xfnm\relax \def\xfnm[#1]{\unskip,\space#1}\fi
\bibitem[{Ausawalaithong et~al.(2018)Ausawalaithong, Thirach, Marukatat and
  Wilaiprasitporn}]{ausawalaithong2018automatic}
\bibinfo{author}{Ausawalaithong, W.}, \bibinfo{author}{Thirach, A.},
  \bibinfo{author}{Marukatat, S.}, \bibinfo{author}{Wilaiprasitporn, T.},
  \bibinfo{year}{2018}.
\newblock \bibinfo{title}{Automatic lung cancer prediction from chest x-ray
  images using the deep learning approach}, in: \bibinfo{booktitle}{2018 11th
  Biomedical Engineering International Conference (BMEiCON)},
  \bibinfo{organization}{IEEE}. pp. \bibinfo{pages}{1--5}.
\bibitem[{Barandela et~al.(2004)Barandela, Valdovinos, S{\'a}nchez and
  Ferri}]{barandela2004imbalanced}
\bibinfo{author}{Barandela, R.}, \bibinfo{author}{Valdovinos, R.M.},
  \bibinfo{author}{S{\'a}nchez, J.S.}, \bibinfo{author}{Ferri, F.J.},
  \bibinfo{year}{2004}.
\newblock \bibinfo{title}{The imbalanced training sample problem: Under or over
  sampling?}, in: \bibinfo{booktitle}{Joint IAPR international workshops on
  statistical techniques in pattern recognition (SPR) and structural and
  syntactic pattern recognition (SSPR)}, \bibinfo{organization}{Springer}. pp.
  \bibinfo{pages}{806--814}.
\bibitem[{Berthelot et~al.(2019)Berthelot, Carlini, Goodfellow, Papernot,
  Oliver and Raffel}]{berthelot2019mixmatch}
\bibinfo{author}{Berthelot, D.}, \bibinfo{author}{Carlini, N.},
  \bibinfo{author}{Goodfellow, I.}, \bibinfo{author}{Papernot, N.},
  \bibinfo{author}{Oliver, A.}, \bibinfo{author}{Raffel, C.A.},
  \bibinfo{year}{2019}.
\newblock \bibinfo{title}{Mixmatch: A holistic approach to semi-supervised
  learning}, in: \bibinfo{booktitle}{Advances in Neural Information Processing
  Systems}, pp. \bibinfo{pages}{5050--5060}.
\bibitem[{Bi et~al.(2017)Bi, Kim, Ahn and Feng}]{bi2017automatic}
\bibinfo{author}{Bi, L.}, \bibinfo{author}{Kim, J.}, \bibinfo{author}{Ahn, E.},
  \bibinfo{author}{Feng, D.}, \bibinfo{year}{2017}.
\newblock \bibinfo{title}{Automatic skin lesion analysis using large-scale
  dermoscopy images and deep residual networks}.
\newblock \bibinfo{journal}{arXiv preprint arXiv:1703.04197} .
\bibitem[{Bunkhumpornpat et~al.(2009)Bunkhumpornpat, Sinapiromsaran and
  Lursinsap}]{bunkhumpornpat2009safe}
\bibinfo{author}{Bunkhumpornpat, C.}, \bibinfo{author}{Sinapiromsaran, K.},
  \bibinfo{author}{Lursinsap, C.}, \bibinfo{year}{2009}.
\newblock \bibinfo{title}{Safe-level-smote: Safe-level-synthetic minority
  over-sampling technique for handling the class imbalanced problem}, in:
  \bibinfo{booktitle}{Pacific-Asia conference on knowledge discovery and data
  mining}, \bibinfo{organization}{Springer}. pp. \bibinfo{pages}{475--482}.
\bibitem[{Chapelle et~al.(2009)Chapelle, Scholkopf and Zien}]{chapelle2009semi}
\bibinfo{author}{Chapelle, O.}, \bibinfo{author}{Scholkopf, B.},
  \bibinfo{author}{Zien, A.}, \bibinfo{year}{2009}.
\newblock \bibinfo{title}{Semi-supervised learning (chapelle, o. et al., eds.;
  2006)[book reviews]}.
\newblock \bibinfo{journal}{IEEE Transactions on Neural Networks}
  \bibinfo{volume}{20}, \bibinfo{pages}{542--542}.
\bibitem[{Chawla et~al.(2002)Chawla, Bowyer, Hall and
  Kegelmeyer}]{chawla2002smote}
\bibinfo{author}{Chawla, N.V.}, \bibinfo{author}{Bowyer, K.W.},
  \bibinfo{author}{Hall, L.O.}, \bibinfo{author}{Kegelmeyer, W.P.},
  \bibinfo{year}{2002}.
\newblock \bibinfo{title}{Smote: synthetic minority over-sampling technique}.
\newblock \bibinfo{journal}{Journal of artificial intelligence research}
  \bibinfo{volume}{16}, \bibinfo{pages}{321--357}.
\bibitem[{Chen et~al.(2006)Chen, Tsai, Moon, Ahn, Young and
  Chen}]{chen2006decision}
\bibinfo{author}{Chen, J.}, \bibinfo{author}{Tsai, C.A.},
  \bibinfo{author}{Moon, H.}, \bibinfo{author}{Ahn, H.},
  \bibinfo{author}{Young, J.}, \bibinfo{author}{Chen, C.H.},
  \bibinfo{year}{2006}.
\newblock \bibinfo{title}{Decision threshold adjustment in class prediction}.
\newblock \bibinfo{journal}{SAR and QSAR in Environmental Research}
  \bibinfo{volume}{17}, \bibinfo{pages}{337--352}.
\bibitem[{Creswell et~al.(2018)Creswell, Pouplin and
  Bharath}]{creswell2018denoising}
\bibinfo{author}{Creswell, A.}, \bibinfo{author}{Pouplin, A.},
  \bibinfo{author}{Bharath, A.A.}, \bibinfo{year}{2018}.
\newblock \bibinfo{title}{Denoising adversarial autoencoders: classifying skin
  lesions using limited labelled training data}.
\newblock \bibinfo{journal}{IET Computer Vision} \bibinfo{volume}{12},
  \bibinfo{pages}{1105--1111}.
\bibitem[{Cubuk et~al.(2019)Cubuk, Zoph, Shlens and Le}]{cubuk2019randaugment}
\bibinfo{author}{Cubuk, E.D.}, \bibinfo{author}{Zoph, B.},
  \bibinfo{author}{Shlens, J.}, \bibinfo{author}{Le, Q.V.},
  \bibinfo{year}{2019}.
\newblock \bibinfo{title}{Randaugment: Practical automated data augmentation
  with a reduced search space}.
\newblock \bibinfo{journal}{arXiv preprint arXiv:1909.13719} .
\bibitem[{Gutman et~al.(2016)Gutman, Codella, Celebi, Helba, Marchetti, Mishra
  and Halpern}]{gutman2016skin}
\bibinfo{author}{Gutman, D.}, \bibinfo{author}{Codella, N.C.},
  \bibinfo{author}{Celebi, E.}, \bibinfo{author}{Helba, B.},
  \bibinfo{author}{Marchetti, M.}, \bibinfo{author}{Mishra, N.},
  \bibinfo{author}{Halpern, A.}, \bibinfo{year}{2016}.
\newblock \bibinfo{title}{Skin lesion analysis toward melanoma detection: A
  challenge at the international symposium on biomedical imaging (isbi) 2016,
  hosted by the international skin imaging collaboration (isic)}.
\newblock \bibinfo{journal}{arXiv preprint arXiv:1605.01397} .
\bibitem[{Han et~al.(2005)Han, Wang and Mao}]{han2005borderline}
\bibinfo{author}{Han, H.}, \bibinfo{author}{Wang, W.Y.}, \bibinfo{author}{Mao,
  B.H.}, \bibinfo{year}{2005}.
\newblock \bibinfo{title}{Borderline-smote: a new over-sampling method in
  imbalanced data sets learning}, in: \bibinfo{booktitle}{International
  conference on intelligent computing}, \bibinfo{organization}{Springer}. pp.
  \bibinfo{pages}{878--887}.
\bibitem[{Han et~al.(2018)Han, Kim, Lim, Park, Park and
  Chang}]{han2018classification}
\bibinfo{author}{Han, S.S.}, \bibinfo{author}{Kim, M.S.}, \bibinfo{author}{Lim,
  W.}, \bibinfo{author}{Park, G.H.}, \bibinfo{author}{Park, I.},
  \bibinfo{author}{Chang, S.E.}, \bibinfo{year}{2018}.
\newblock \bibinfo{title}{Classification of the clinical images for benign and
  malignant cutaneous tumors using a deep learning algorithm}.
\newblock \bibinfo{journal}{Journal of Investigative Dermatology}
  \bibinfo{volume}{138}, \bibinfo{pages}{1529--1538}.
\bibitem[{Hekler et~al.(2019)Hekler, Utikal, Enk, Hauschild, Weichenthal,
  Maron, Berking, Haferkamp, Klode, Schadendorf et~al.}]{hekler2019superior}
\bibinfo{author}{Hekler, A.}, \bibinfo{author}{Utikal, J.S.},
  \bibinfo{author}{Enk, A.H.}, \bibinfo{author}{Hauschild, A.},
  \bibinfo{author}{Weichenthal, M.}, \bibinfo{author}{Maron, R.C.},
  \bibinfo{author}{Berking, C.}, \bibinfo{author}{Haferkamp, S.},
  \bibinfo{author}{Klode, J.}, \bibinfo{author}{Schadendorf, D.}, et~al.,
  \bibinfo{year}{2019}.
\newblock \bibinfo{title}{Superior skin cancer classification by the
  combination of human and artificial intelligence}.
\newblock \bibinfo{journal}{European Journal of Cancer} \bibinfo{volume}{120},
  \bibinfo{pages}{114--121}.
\bibitem[{Hyun et~al.(2020)Hyun, Jeong and Kwak}]{hyun2020class}
\bibinfo{author}{Hyun, M.}, \bibinfo{author}{Jeong, J.}, \bibinfo{author}{Kwak,
  N.}, \bibinfo{year}{2020}.
\newblock \bibinfo{title}{Class-imbalanced semi-supervised learning}.
\newblock \bibinfo{journal}{arXiv preprint arXiv:2002.06815} .
\bibitem[{Iizuka et~al.(2020)Iizuka, Kanavati, Kato, Rambeau, Arihiro and
  Tsuneki}]{iizuka2020deep}
\bibinfo{author}{Iizuka, O.}, \bibinfo{author}{Kanavati, F.},
  \bibinfo{author}{Kato, K.}, \bibinfo{author}{Rambeau, M.},
  \bibinfo{author}{Arihiro, K.}, \bibinfo{author}{Tsuneki, M.},
  \bibinfo{year}{2020}.
\newblock \bibinfo{title}{Deep learning models for histopathological
  classification of gastric and colonic epithelial tumours}.
\newblock \bibinfo{journal}{Scientific Reports} \bibinfo{volume}{10},
  \bibinfo{pages}{1--11}.
\bibitem[{Kaur et~al.(2019)Kaur, Pannu and Malhi}]{kaur2019systematic}
\bibinfo{author}{Kaur, H.}, \bibinfo{author}{Pannu, H.S.},
  \bibinfo{author}{Malhi, A.K.}, \bibinfo{year}{2019}.
\newblock \bibinfo{title}{A systematic review on imbalanced data challenges in
  machine learning: Applications and solutions}.
\newblock \bibinfo{journal}{ACM Computing Surveys (CSUR)} \bibinfo{volume}{52},
  \bibinfo{pages}{1--36}.
\bibitem[{Khan et~al.(2017)Khan, Hayat, Bennamoun, Sohel and
  Togneri}]{khan2017cost}
\bibinfo{author}{Khan, S.H.}, \bibinfo{author}{Hayat, M.},
  \bibinfo{author}{Bennamoun, M.}, \bibinfo{author}{Sohel, F.A.},
  \bibinfo{author}{Togneri, R.}, \bibinfo{year}{2017}.
\newblock \bibinfo{title}{Cost-sensitive learning of deep feature
  representations from imbalanced data}.
\newblock \bibinfo{journal}{IEEE transactions on neural networks and learning
  systems} \bibinfo{volume}{29}, \bibinfo{pages}{3573--3587}.
\bibitem[{Krawczyk and Wo{\'z}niak(2015)}]{krawczyk2015cost}
\bibinfo{author}{Krawczyk, B.}, \bibinfo{author}{Wo{\'z}niak, M.},
  \bibinfo{year}{2015}.
\newblock \bibinfo{title}{Cost-sensitive neural network with roc-based moving
  threshold for imbalanced classification}, in:
  \bibinfo{booktitle}{International Conference on Intelligent Data Engineering
  and Automated Learning}, \bibinfo{organization}{Springer}. pp.
  \bibinfo{pages}{45--52}.
\bibitem[{Krizhevsky et~al.(2012)Krizhevsky, Sutskever and
  Hinton}]{krizhevsky2012imagenet}
\bibinfo{author}{Krizhevsky, A.}, \bibinfo{author}{Sutskever, I.},
  \bibinfo{author}{Hinton, G.E.}, \bibinfo{year}{2012}.
\newblock \bibinfo{title}{Imagenet classification with deep convolutional
  neural networks}, in: \bibinfo{booktitle}{Advances in neural information
  processing systems}, pp. \bibinfo{pages}{1097--1105}.
\bibitem[{Kubat et~al.(1997)Kubat, Matwin et~al.}]{kubat1997addressing}
\bibinfo{author}{Kubat, M.}, \bibinfo{author}{Matwin, S.}, et~al.,
  \bibinfo{year}{1997}.
\newblock \bibinfo{title}{Addressing the curse of imbalanced training sets:
  one-sided selection}, in: \bibinfo{booktitle}{Icml},
  \bibinfo{organization}{Citeseer}. pp. \bibinfo{pages}{179--186}.
\bibitem[{Laine and Aila(2016)}]{laine2016temporal}
\bibinfo{author}{Laine, S.}, \bibinfo{author}{Aila, T.}, \bibinfo{year}{2016}.
\newblock \bibinfo{title}{Temporal ensembling for semi-supervised learning}.
\newblock \bibinfo{journal}{arXiv preprint arXiv:1610.02242} .
\bibitem[{Lin et~al.(2017)Lin, Goyal, Girshick, He and
  Doll{\'a}r}]{lin2017focal}
\bibinfo{author}{Lin, T.Y.}, \bibinfo{author}{Goyal, P.},
  \bibinfo{author}{Girshick, R.}, \bibinfo{author}{He, K.},
  \bibinfo{author}{Doll{\'a}r, P.}, \bibinfo{year}{2017}.
\newblock \bibinfo{title}{Focal loss for dense object detection}, in:
  \bibinfo{booktitle}{Proceedings of the IEEE international conference on
  computer vision}, pp. \bibinfo{pages}{2980--2988}.
\bibitem[{Mani and Zhang(2003)}]{mani2003knn}
\bibinfo{author}{Mani, I.}, \bibinfo{author}{Zhang, I.}, \bibinfo{year}{2003}.
\newblock \bibinfo{title}{knn approach to unbalanced data distributions: a case
  study involving information extraction}, in: \bibinfo{booktitle}{Proceedings
  of workshop on learning from imbalanced datasets}.
\bibitem[{Miyato et~al.(2018)Miyato, Maeda, Koyama and
  Ishii}]{miyato2018virtual}
\bibinfo{author}{Miyato, T.}, \bibinfo{author}{Maeda, S.i.},
  \bibinfo{author}{Koyama, M.}, \bibinfo{author}{Ishii, S.},
  \bibinfo{year}{2018}.
\newblock \bibinfo{title}{Virtual adversarial training: a regularization method
  for supervised and semi-supervised learning}.
\newblock \bibinfo{journal}{IEEE transactions on pattern analysis and machine
  intelligence} \bibinfo{volume}{41}, \bibinfo{pages}{1979--1993}.
\bibitem[{Orlando et~al.(2020)Orlando, Fu, Breda, van Keer, Bathula,
  Diaz-Pinto, Fang, Heng, Kim, Lee et~al.}]{orlando2020refuge}
\bibinfo{author}{Orlando, J.I.}, \bibinfo{author}{Fu, H.},
  \bibinfo{author}{Breda, J.B.}, \bibinfo{author}{van Keer, K.},
  \bibinfo{author}{Bathula, D.R.}, \bibinfo{author}{Diaz-Pinto, A.},
  \bibinfo{author}{Fang, R.}, \bibinfo{author}{Heng, P.A.},
  \bibinfo{author}{Kim, J.}, \bibinfo{author}{Lee, J.}, et~al.,
  \bibinfo{year}{2020}.
\newblock \bibinfo{title}{Refuge challenge: A unified framework for evaluating
  automated methods for glaucoma assessment from fundus photographs}.
\newblock \bibinfo{journal}{Medical image analysis} \bibinfo{volume}{59},
  \bibinfo{pages}{101570}.
\bibitem[{Phan et~al.(2017)Phan, Krawczyk-Becker, Gerkmann and
  Mertins}]{phan2017dnn}
\bibinfo{author}{Phan, H.}, \bibinfo{author}{Krawczyk-Becker, M.},
  \bibinfo{author}{Gerkmann, T.}, \bibinfo{author}{Mertins, A.},
  \bibinfo{year}{2017}.
\newblock \bibinfo{title}{Dnn and cnn with weighted and multi-task loss
  functions for audio event detection}.
\newblock \bibinfo{journal}{arXiv preprint arXiv:1708.03211} .
\bibitem[{Rajpurkar et~al.(2017)Rajpurkar, Irvin, Zhu, Yang, Mehta, Duan, Ding,
  Bagul, Langlotz, Shpanskaya et~al.}]{rajpurkar2017chexnet}
\bibinfo{author}{Rajpurkar, P.}, \bibinfo{author}{Irvin, J.},
  \bibinfo{author}{Zhu, K.}, \bibinfo{author}{Yang, B.},
  \bibinfo{author}{Mehta, H.}, \bibinfo{author}{Duan, T.},
  \bibinfo{author}{Ding, D.}, \bibinfo{author}{Bagul, A.},
  \bibinfo{author}{Langlotz, C.}, \bibinfo{author}{Shpanskaya, K.}, et~al.,
  \bibinfo{year}{2017}.
\newblock \bibinfo{title}{Chexnet: Radiologist-level pneumonia detection on
  chest x-rays with deep learning}.
\newblock \bibinfo{journal}{arXiv preprint arXiv:1711.05225} .
\bibitem[{Rezvantalab et~al.(2018)Rezvantalab, Safigholi and
  Karimijeshni}]{rezvantalab2018dermatologist}
\bibinfo{author}{Rezvantalab, A.}, \bibinfo{author}{Safigholi, H.},
  \bibinfo{author}{Karimijeshni, S.}, \bibinfo{year}{2018}.
\newblock \bibinfo{title}{Dermatologist level dermoscopy skin cancer
  classification using different deep learning convolutional neural networks
  algorithms}.
\newblock \bibinfo{journal}{arXiv preprint arXiv:1810.10348} .
\bibitem[{Springenberg(2015)}]{springenberg2015unsupervised}
\bibinfo{author}{Springenberg, J.T.}, \bibinfo{year}{2015}.
\newblock \bibinfo{title}{Unsupervised and semi-supervised learning with
  categorical generative adversarial networks}.
\newblock \bibinfo{journal}{arXiv preprint arXiv:1511.06390} .
\bibitem[{Su et~al.(2019)Su, Shi, Cai and Yang}]{su2019local}
\bibinfo{author}{Su, H.}, \bibinfo{author}{Shi, X.}, \bibinfo{author}{Cai, J.},
  \bibinfo{author}{Yang, L.}, \bibinfo{year}{2019}.
\newblock \bibinfo{title}{Local and global consistency regularized mean teacher
  for semi-supervised nuclei classification}, in:
  \bibinfo{booktitle}{International Conference on Medical Image Computing and
  Computer-Assisted Intervention}, \bibinfo{organization}{Springer}. pp.
  \bibinfo{pages}{559--567}.
\bibitem[{Tarvainen and Valpola(2017)}]{tarvainen2017mean}
\bibinfo{author}{Tarvainen, A.}, \bibinfo{author}{Valpola, H.},
  \bibinfo{year}{2017}.
\newblock \bibinfo{title}{Mean teachers are better role models: Weight-averaged
  consistency targets improve semi-supervised deep learning results}, in:
  \bibinfo{booktitle}{Advances in neural information processing systems}, pp.
  \bibinfo{pages}{1195--1204}.
\bibitem[{Tschandl et~al.(2018)Tschandl, Rosendahl and
  Kittler}]{tschandl2018ham10000}
\bibinfo{author}{Tschandl, P.}, \bibinfo{author}{Rosendahl, C.},
  \bibinfo{author}{Kittler, H.}, \bibinfo{year}{2018}.
\newblock \bibinfo{title}{The ham10000 dataset, a large collection of
  multi-source dermatoscopic images of common pigmented skin lesions}.
\newblock \bibinfo{journal}{Scientific data} \bibinfo{volume}{5},
  \bibinfo{pages}{180161}.
\bibitem[{Wang et~al.(2018)Wang, Cui, Chen, Avidan, Abdallah and
  Kronzer}]{wang2018predicting}
\bibinfo{author}{Wang, H.}, \bibinfo{author}{Cui, Z.}, \bibinfo{author}{Chen,
  Y.}, \bibinfo{author}{Avidan, M.}, \bibinfo{author}{Abdallah, A.B.},
  \bibinfo{author}{Kronzer, A.}, \bibinfo{year}{2018}.
\newblock \bibinfo{title}{Predicting hospital readmission via cost-sensitive
  deep learning}.
\newblock \bibinfo{journal}{IEEE/ACM transactions on computational biology and
  bioinformatics} \bibinfo{volume}{15}, \bibinfo{pages}{1968--1978}.
\bibitem[{Wang et~al.(2016)Wang, Liu, Wu, Cao, Meng and
  Kennedy}]{wang2016training}
\bibinfo{author}{Wang, S.}, \bibinfo{author}{Liu, W.}, \bibinfo{author}{Wu,
  J.}, \bibinfo{author}{Cao, L.}, \bibinfo{author}{Meng, Q.},
  \bibinfo{author}{Kennedy, P.J.}, \bibinfo{year}{2016}.
\newblock \bibinfo{title}{Training deep neural networks on imbalanced data
  sets}, in: \bibinfo{booktitle}{2016 international joint conference on neural
  networks (IJCNN)}, \bibinfo{organization}{IEEE}. pp.
  \bibinfo{pages}{4368--4374}.
\bibitem[{Xie et~al.(2019)Xie, Dai, Hovy, Luong and Le}]{xie2019unsupervised}
\bibinfo{author}{Xie, Q.}, \bibinfo{author}{Dai, Z.}, \bibinfo{author}{Hovy,
  E.}, \bibinfo{author}{Luong, M.T.}, \bibinfo{author}{Le, Q.V.},
  \bibinfo{year}{2019}.
\newblock \bibinfo{title}{Unsupervised data augmentation}.
\newblock \bibinfo{journal}{arXiv preprint arXiv:1904.12848} .
\bibitem[{Yang et~al.(2019)Yang, Zhao, Qiang, Yang, Dong, Du, Shi and
  Zia}]{yang2019dscgans}
\bibinfo{author}{Yang, W.}, \bibinfo{author}{Zhao, J.}, \bibinfo{author}{Qiang,
  Y.}, \bibinfo{author}{Yang, X.}, \bibinfo{author}{Dong, Y.},
  \bibinfo{author}{Du, Q.}, \bibinfo{author}{Shi, G.}, \bibinfo{author}{Zia,
  M.B.}, \bibinfo{year}{2019}.
\newblock \bibinfo{title}{Dscgans: Integrate domain knowledge in training
  dual-path semi-supervised conditional generative adversarial networks and
  s3vm for ultrasonography thyroid nodules classification}, in:
  \bibinfo{booktitle}{International Conference on Medical Image Computing and
  Computer-Assisted Intervention}, \bibinfo{organization}{Springer}. pp.
  \bibinfo{pages}{558--566}.
\bibitem[{Yu et~al.(2016)Yu, Sun, Yang, Yang, Shen and Qi}]{yu2016odoc}
\bibinfo{author}{Yu, H.}, \bibinfo{author}{Sun, C.}, \bibinfo{author}{Yang,
  X.}, \bibinfo{author}{Yang, W.}, \bibinfo{author}{Shen, J.},
  \bibinfo{author}{Qi, Y.}, \bibinfo{year}{2016}.
\newblock \bibinfo{title}{Odoc-elm: Optimal decision outputs compensation-based
  extreme learning machine for classifying imbalanced data}.
\newblock \bibinfo{journal}{Knowledge-Based Systems} \bibinfo{volume}{92},
  \bibinfo{pages}{55--70}.
\bibitem[{Zhang et~al.(2016)Zhang, Tan and Ren}]{zhang2016training}
\bibinfo{author}{Zhang, C.}, \bibinfo{author}{Tan, K.C.}, \bibinfo{author}{Ren,
  R.}, \bibinfo{year}{2016}.
\newblock \bibinfo{title}{Training cost-sensitive deep belief networks on
  imbalance data problems}, in: \bibinfo{booktitle}{2016 international joint
  conference on neural networks (IJCNN)}, \bibinfo{organization}{IEEE}. pp.
  \bibinfo{pages}{4362--4367}.
\bibitem[{Zhong et~al.(2017)Zhong, Zheng, Kang, Li and Yang}]{zhong2017random}
\bibinfo{author}{Zhong, Z.}, \bibinfo{author}{Zheng, L.},
  \bibinfo{author}{Kang, G.}, \bibinfo{author}{Li, S.}, \bibinfo{author}{Yang,
  Y.}, \bibinfo{year}{2017}.
\newblock \bibinfo{title}{Random erasing data augmentation}.
\newblock \bibinfo{journal}{arXiv preprint arXiv:1708.04896} .

\end{thebibliography}
\end{document}